%% file: opsd_v.tex
\PassOptionsToPackage{table}{xcolor}
\documentclass{article}

\usepackage{opsdv_preprint,times}
\usepackage[letterpaper,left=0.9in,right=0.9in,top=0.68in,bottom=0.72in]{geometry}
\usepackage{titlesec}

\input{math_commands.tex}

\usepackage[utf8]{inputenc}
\usepackage[T1]{fontenc}
\usepackage{amsmath}
\usepackage{amssymb}
\usepackage{booktabs}
\usepackage{enumitem}
\usepackage{graphicx}
\usepackage{hyperref}
\usepackage{url}
\usepackage{algorithm}
\usepackage{algorithmic}
\usepackage[capitalize,nameinlink]{cleveref}
\usepackage{multirow}
\usepackage{xcolor}
\usepackage{tikz}

\definecolor{headerblue}{HTML}{EAF2FF}
\definecolor{oursblue}{HTML}{F3F8FF}
\definecolor{bestgreen}{HTML}{EAF7EA}
\definecolor{opsdgreen}{HTML}{174B3B}
\definecolor{frontgray}{HTML}{F1F2F3}
\definecolor{frontdark}{HTML}{4B4B4B}
\definecolor{studybase}{HTML}{3B6F8F}
\definecolor{studysame}{HTML}{9A9A9A}
\newcommand{\best}[1]{\cellcolor{bestgreen}\textbf{#1}}
\usepackage{xspace}
\usepackage{tcolorbox}
\usepackage{caption}
\usepackage{placeins}
\usepackage{float}
\graphicspath{{fig/}}

\hypersetup{
    colorlinks=true,
    linkcolor=blue,
    urlcolor=blue,
    citecolor=blue
}

\newcommand{\method}{\texorpdfstring{\textcolor{opsdgreen}{OPSD-V}}{OPSD-V}\xspace}

\newtcolorbox{figureplaceholder}[1]{
    colback=gray!6!white,
    colframe=gray!55!black,
    boxrule=0.8pt,
    arc=3pt,
    left=8pt,
    right=8pt,
    top=8pt,
    bottom=8pt,
    title=#1
}
\newtcolorbox{motivationbox}{
    colback=opsdgreen!8!white,
    colframe=opsdgreen!28!white,
    boxrule=0pt,
    arc=4pt,
    left=8pt,
    right=8pt,
    top=6pt,
    bottom=6pt,
    before skip=8pt,
    after skip=8pt
}
\newtcolorbox{frontmatterbox}{
    colback=frontgray,
    colframe=frontgray,
    boxrule=0pt,
    arc=8pt,
    left=20pt,
    right=20pt,
    top=18pt,
    bottom=18pt,
    width=\textwidth,
    before skip=0pt,
    after skip=0pt
}

\titleformat{\section}{\sffamily\Large\bfseries}{\thesection}{1em}{}
\titleformat{\subsection}{\sffamily\large\bfseries}{\thesubsection}{0.8em}{}
\titleformat{\subsubsection}{\sffamily\normalsize\bfseries}{\thesubsubsection}{0.8em}{}
\titlespacing*{\section}{0pt}{2.0ex plus 0.4ex minus 0.2ex}{1.1ex}
\titlespacing*{\subsection}{0pt}{1.7ex plus 0.3ex minus 0.2ex}{0.8ex}
\titlespacing*{\subsubsection}{0pt}{1.4ex plus 0.3ex minus 0.2ex}{0.6ex}

\newcommand{\papersubtitle}{On-Policy Self-Distillation for Post-Training Few-Step Autoregressive Video Generators}
\newcommand{\paperfrontsubtitle}{On-Policy Self-Distillation for Post-Training Few-Step\\Autoregressive Video Generators}
\newcommand{\projectpageurl}{https://meigen-ai.github.io/OPSD-V}
\newcommand{\codeurl}{https://github.com/MeiGen-AI/OPSD-V}
\newcommand{\codelabel}{MeiGen-AI/OPSD-V}

\title{\method: \papersubtitle}

\author{
\textbf{Hongyu Liu\textsuperscript{1,2},
Chun Wang\textsuperscript{1,2},
Feng Gao\textsuperscript{1,$\dagger$},
Xuanhua He\textsuperscript{1,2},} \\
\textbf{Yue Ma\textsuperscript{2},
Ziyu Wan\textsuperscript{3},
Yong Zhang\textsuperscript{1,$\ddagger$},
Xiaoming Wei\textsuperscript{1},
Qifeng Chen\textsuperscript{2,$\dagger$}} \\
\textsuperscript{1}Meituan \quad
\textsuperscript{2}HKUST \quad
\textsuperscript{3}City University of Hong Kong \\
\textsuperscript{$\dagger$}Corresponding authors \quad
\textsuperscript{$\ddagger$}Project lead \\
\texttt{hliudq@connect.ust.hk}
}

\newcommand{\frontauthorblock}{%
\textbf{Hongyu Liu\textsuperscript{1,2},
Chun Wang\textsuperscript{1,2},
Feng Gao\textsuperscript{1,$\dagger$},
Xuanhua He\textsuperscript{1,2},}\\
\textbf{Yue Ma\textsuperscript{2},
Ziyu Wan\textsuperscript{3},
Yong Zhang\textsuperscript{1,$\ddagger$},
Xiaoming Wei\textsuperscript{1},
Qifeng Chen\textsuperscript{2,$\dagger$}}\\
\textsuperscript{1}Meituan \quad
\textsuperscript{2}HKUST \quad
\textsuperscript{3}City University of Hong Kong\\
\textsuperscript{$\dagger$}Corresponding authors \quad
\textsuperscript{$\ddagger$}Project lead\\
\texttt{hliudq@connect.ust.hk}%
}

\newcommand{\frontaffiliationlogos}{%
\includegraphics[height=0.18in]{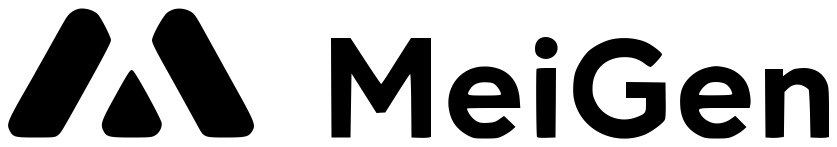}\hspace{0.10in}%
\includegraphics[height=0.23in]{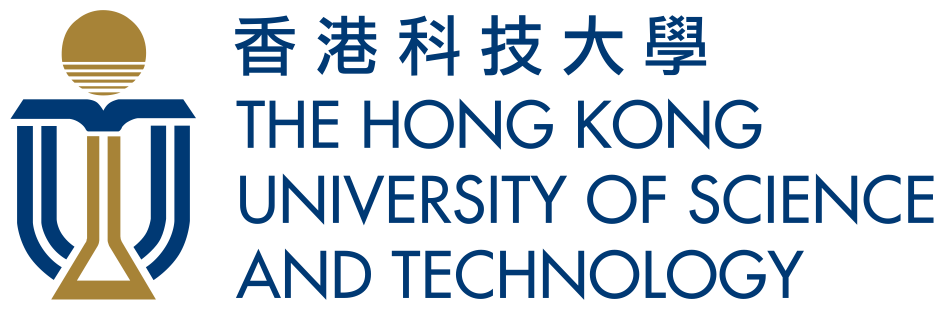}\hspace{0.10in}%
\includegraphics[height=0.25in]{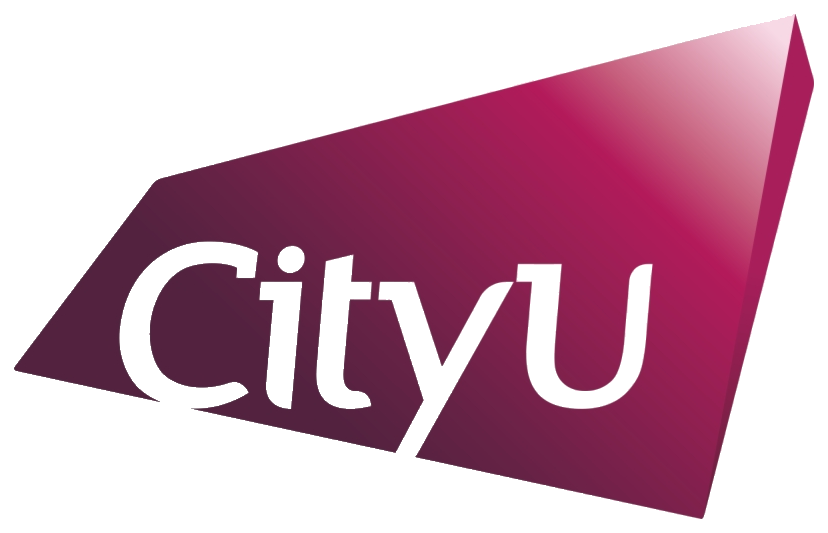}%
}

\newcommand{\makefrontmatter}{%
\vspace*{0.16in}
\begin{frontmatterbox}
{\sffamily\bfseries\fontsize{28}{31}\selectfont \textcolor{opsdgreen}{OPSD-V}\par}
\vspace{0.08in}
{\sffamily\bfseries\Large\color{frontdark}\paperfrontsubtitle\par}
\vspace{0.22in}
{\normalsize\frontauthorblock\par}
\vspace{0.26in}
{\normalsize\paperabstract\par}
\vspace{0.22in}
\noindent
\begin{minipage}[b]{0.60\linewidth}
{\small
\textbf{Project page:} \href{\projectpageurl}{\texttt{\projectpageurl}}\\
\textbf{Code:} \href{\codeurl}{\texttt{\codelabel}}%
}
\end{minipage}%
\hfill
\begin{minipage}[b]{0.34\linewidth}
\raggedleft
\resizebox{\linewidth}{!}{\frontaffiliationlogos}
\end{minipage}
\end{frontmatterbox}
\vspace{0.22in}
}

\arxivcopy 
\renewcommand{\headrulewidth}{0pt}
\renewcommand{\footrulewidth}{0pt}
\AddToShipoutPicture{%
  \AtPageLowerLeft{%
    \raisebox{0.34in}{\makebox[\paperwidth][c]{\thepage}}%
  }%
}

\input{section/abstract}

\begin{document}
\makefrontmatter
\input{section/introduction}
\input{section/related_work}

\input{section/preliminaries}

\input{section/method}

\input{section/experiments}

\input{section/analysis}
\input{section/conclusion}

\input{section/acknowledgements}

\bibliographystyle{opsdv_preprint}
\bibliography{references}

\end{document}

%% file: math_commands.tex
\usepackage{amsmath,amsfonts,bm}

\def\eqref#1{equation~\ref{#1}}

\def\1{\bm{1}}

\DeclareMathAlphabet{\mathsfit}{\encodingdefault}{\sfdefault}{m}{sl}
\SetMathAlphabet{\mathsfit}{bold}{\encodingdefault}{\sfdefault}{bx}{n}



%% file: section/abstract.tex
\newcommand{\paperabstract}{%
We propose \method, an on-policy self-distillation paradigm for post-training few-step autoregressive (AR) video diffusion models. Existing few-step AR video generators, often obtained through DMD-style distillation, can generate long videos with low latency, but still suffer from error accumulation and weakened motion dynamics during long autoregressive rollout. \method aims to further reduce long-horizon degradation and improve motion dynamics while preserving the original few-step inference path. Our key idea is to introduce real long-video data as temporal context during training and use it to provide dense trajectory-level supervision. Compared with relying only on a short-clip teacher distribution, real long videos offer a richer and cleaner target distribution for supervising long AR rollouts. Specifically, the student follows the exact inference-time rollout, generating each chunk conditioned on its own previously generated KV cache. In parallel, the teacher is evaluated at the same student-visited denoising states, but uses a cleaner AR-consistent temporal cache in which older history can be replaced by real-video context. To maintain autoregressive consistency and prevent the teacher from becoming a fully teacher-forced oracle, both branches share an initial real-video prefix, and the teacher keeps its most recent cache chunk generated by the model itself. This design provides dense denoising-level corrective targets under on-policy AR cache dynamics, without changing the sampler, number of denoising steps, or inference-time cache mechanism. We apply \method to representative few-step AR video models, including Self-Forcing and LongLive. Experiments show consistent improvements in visual quality, motion dynamics, and VBenchLong scores. In a user study with 10 participants comparing 20 video pairs, \method is preferred over the base models in 66.0\% of overall-preference judgments (82.5\% excluding ties), demonstrating the effectiveness of on-policy self-distillation with real long-video context for long-horizon AR video generation.
}

%% file: section/introduction.tex
\section{Introduction}


Video generation has rapidly evolved from short, offline clip synthesis to large-scale video foundation models capable of producing high-resolution, text-aligned, and temporally coherent videos. Modern video generation models are largely built upon diffusion transformers (DiTs)~\citep{Peebles2022DiT}, whose scalability has enabled substantial progress in visual fidelity, motion quality, prompt following, and long-video generation~\citep{cogvideox2024,hunyuanvideo2024,wan2025,longcatvideo2025,hacohen2026ltx,videoworldsimulators2024}. Despite these advances, most large-scale video foundation models are still primarily designed for offline generation, where an entire video is synthesized after sampling rather than produced continuously during user interaction.

In contrast to offline video generation, real-time video generation requires a model to produce content sequentially with low viewing latency. This requirement naturally favors autoregressive (AR) video models, where each new frame or chunk is generated conditioned on previous outputs, and transformer KV caches are reused to maintain historical context efficiently. To make AR generation practical, recent methods combine causal video modeling with few-step distillation, which compresses slow diffusion or flow models into efficient few-step generators~\citep{yin2024one,yin2024improved,wang2023prolificdreamer}. Building on this recipe, CausVid distills a bidirectional video DiT into a causal AR student for fast streaming generation~\citep{causvid2024}, while Self-Forcing further aligns training with inference by rolling out the model on its own generated frames and rolling KV cache~\citep{selfforcing2025}. Together, these AR and few-step generation techniques are enabling applications such as streaming avatars~\citep{yang2025infinitetalkaudiodrivenvideogeneration,team2026longcat,wang2026flowact}, interactive media~\citep{shin2025motionstream,ai2025magi1autoregressivevideogeneration}, world-model-like applications~\citep{genie3,lingbot-world}, and embodied intelligence~\citep{lingbot-va2026,ye2026worldactionmodelszeroshot}, where models must produce videos continuously while maintaining stable identities, motions, and scene structures over long horizons.
 
\begin{figure}[t]
\centering
  \includegraphics[width=\linewidth]{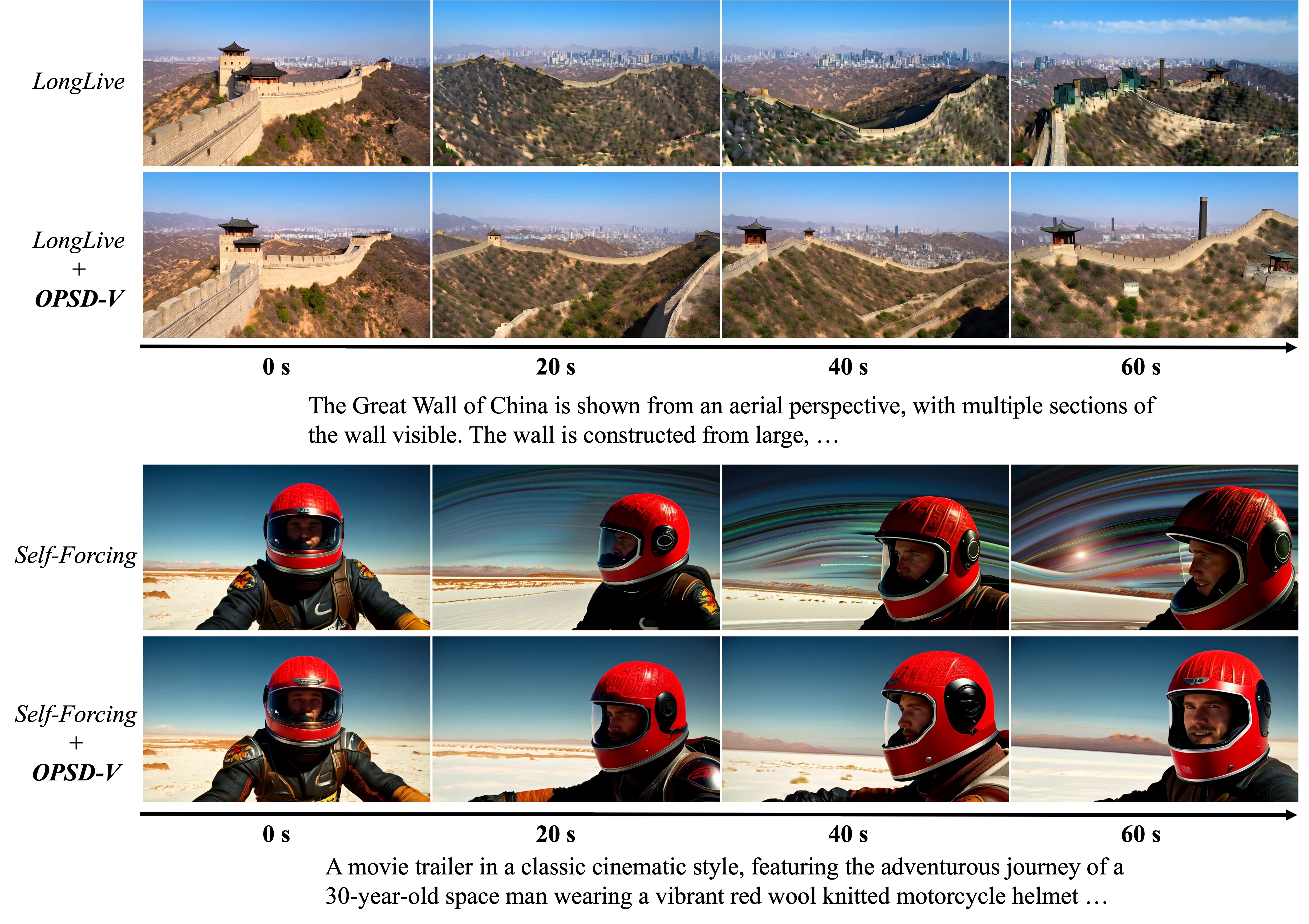}
\caption{
\textbf{\method improves long-horizon autoregressive video generation.}
We compare the original base model and its \method post-trained version under the same prompt and seed. The first example is based on LongLive, and the second example is based on Self-Forcing. In each example, the first row shows the base model result, while the second row shows the result after applying \method. Our method produces more dynamic and temporally stable videos while reducing long-horizon error accumulation. For LongLive, \method noticeably reduces artifacts in the grass region and distant background. For Self-Forcing, \method improves the character's head motion and suppresses background artifacts.
}
  \label{fig:teaser}
\end{figure}

Despite the success of combining self-forcing style training with DMD-style few-step distillation, existing AR video generators still suffer from long-horizon degradation. One fundamental limitation is that the target distribution used in DMD-style training is usually provided by a bidirectional short-clip video teacher, which cannot directly supervise truly long autoregressive trajectories. Although recent methods~\cite{longlive2025, selfforcingpp2025, rewardforcing2026, rollingsink2026} extend rollout length and adopt attention-sink mechanisms~\cite{xiao2023streamingllm} to mitigate error accumulation, their supervision is still limited by the capability and temporal range of the underlying clip-level teacher. Moreover, DMD itself may introduce side effects such as weakened dynamics and color drift, and its supervision is typically defined at the chunk or clip distribution level rather than as dense corrective targets along the entire denoising trajectory.

This limitation points to an inherent ceiling of existing DMD-style post-training: the few-step sampling path is preserved, but the score-matching signal is still tied to the temporal range and quality of a finite short-clip teacher. This raises a central question for post-training few-step AR video generators:
\begin{motivationbox}
\centering\itshape
Can real long-video data serve as stronger training supervision while preserving the original few-step AR generation capability?
\end{motivationbox}
Our answer is to use real long videos not as direct teacher-forcing targets, but as privileged temporal context for constructing a cleaner teacher distribution on the student's own rollout states. On-policy self-distillation (OPSD) provides a suitable framework for this goal. Recently studied in autoregressive large language models~\citep{selfdistilledreasoner2026,sdft2026,sdpo2026}, OPSD adopts an on-policy learning paradigm in which the model samples from its current policy as a student, while a stronger teacher distribution is obtained by conditioning the same model on richer in-context information. This formulation naturally provides dense supervision along the model's own trajectory. In our AR video setting, real long videos can serve as such richer context: by conditioning the same video generator on cleaner long-video states, the teacher can provide a more reliable target distribution for the student's rollout trajectory. A closely related work, D-OPSD~\citep{dopsd2026}, has demonstrated the feasibility of this context-enhanced self-distillation idea for text-to-image generation, where a real image paired with the text prompt is used as additional context to obtain a stronger teacher distribution. Inspired by this paradigm, we explore using real long videos as reliable temporal context to construct a cleaner teacher distribution, enabling dense denoising-level supervision on the student's autoregressive rollout states.

\begin{figure}[!t]
\centering
\includegraphics[width=\linewidth]{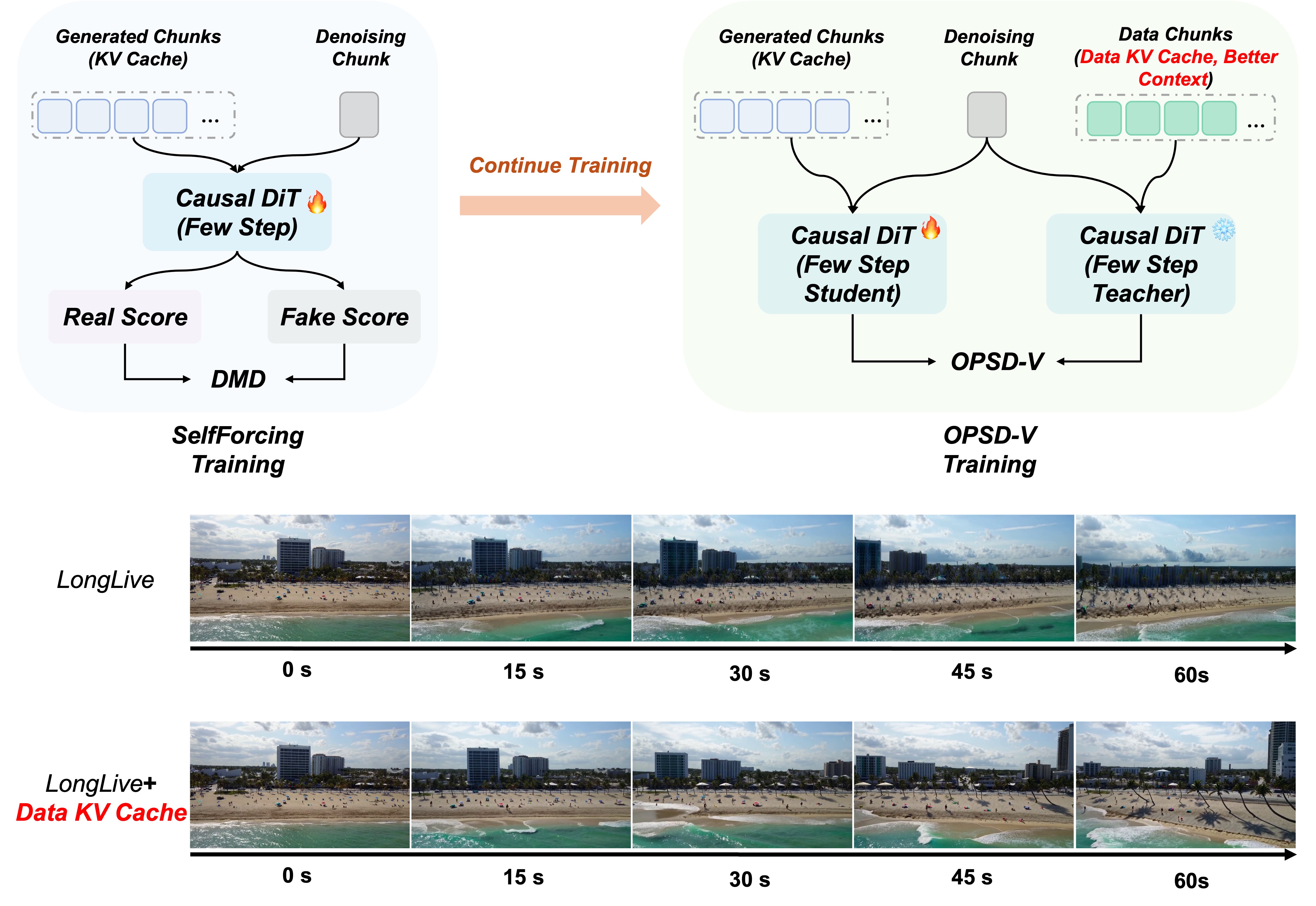}
\caption{\textbf{Overview and motivating cache intervention.} \emph{Top:} Unlike Self-Forcing training, which applies DMD supervision after self-rollout, \method continues post-training an existing few-step AR generator by aligning an on-policy student with a teacher conditioned on cleaner long-video context. The student writes its own generated chunks into the KV cache, while the teacher is evaluated at the same denoising states using a data-assisted cache. \emph{Bottom:} Before any \method post-training, we isolate the effect of cache quality using the original LongLive model at test time. Both rows are initialized with the same real first chunk from the source video. The first row is the standard LongLive rollout; in the second row, older KV-cache entries are replaced by those computed from the corresponding real-video chunks, while the most recent cache chunk remains generated by the model. This data-context intervention already improves long-horizon stability, motivating our use of real-video context to construct cleaner teacher targets during training.}
\label{fig:overview}
\end{figure}

In this paper, we propose \method, a cache-aware on-policy self-distillation framework for post-training few-step AR video diffusion models. As illustrated in Fig.~\ref{fig:overview}, \method starts from an existing few-step AR generator and continues training it with supervision tailored to long autoregressive rollout. A key challenge is that directly using real long videos is non-trivial: even under the same prompt, the student rollout may produce a trajectory different from the real video, so naively using real-video chunks as reconstruction targets or teacher context can introduce semantic and temporal mismatch. Therefore, we treat real videos as privileged temporal context rather than direct output targets. Specifically, the student remains fully on-policy: it denoises each chunk with the original few-step sampler, writes its own generated chunks into the KV cache, and continues from the resulting self-induced temporal states. The teacher is evaluated at the same temporal positions, timesteps, and student-visited noisy latents, but uses a cleaner cache constructed from real long-video context. Both branches share the same initial real-video prefix to anchor the scene; for each target chunk, the teacher replaces older generated cache history with the corresponding real-video context while retaining the most recent generated chunk. This design reduces accumulated degradation in the teacher context while preserving autoregressive continuation from the student's current state. Dense velocity matching along the fixed few-step denoising trajectory then improves long-horizon generation without changing the original low-latency sampling path.

The lower part of Fig.~\ref{fig:overview} shows the diagnostic observation that led to this design. Before applying any \method post-training, we take the original LongLive model and run it at test time from the same real first chunk under two cache settings. In both settings, this first chunk is taken from the source video and used only to initialize the cache. The first setting is standard inference, where the history cache is built entirely from generated chunks. In the second setting, we provide data context only to the older cache history by replacing those KV entries with features computed from the corresponding real-video chunks, while keeping the most recent cache chunk generated by the model itself. No model parameters are updated, and the denoising sampler is unchanged. The fact that this test-time data-context intervention alone produces a visibly more stable long-horizon rollout directly suggests that degradation in the generated KV cache is a key bottleneck. This observation motivates \method: during training, we use real long-video context as privileged teacher information to provide cleaner targets, while the student still follows the same self-generated cache trajectory it will encounter at deployment.

We perform \method post-training on a small customized long-video dataset containing 3,800 videos, each approximately one minute in length. To verify the generality of our post-training framework, we apply \method to two representative few-step AR video generators, Self-Forcing~\cite{selfforcing2025} and LongLive~\cite{longlive2025}, using LoRA-based continued training for both models. \textbf{Self-Forcing is a widely compared few-step AR baseline}, but its original training is mainly performed on short clips; using it as the only long-video baseline would make the evaluation incomplete and could overstate gains from long-horizon post-training. Since \method adopts an attention-sink cache mechanism by default for long-video training and inference, we additionally equip Self-Forcing with the same attention-sink mechanism in all long-video evaluations to make the comparison stronger. \textbf{To further ensure fairness and validate \method on a backbone already trained for streaming long-video generation}, we also train and compare on LongLive, which incorporates streaming long tuning for long-video synthesis. After post-training, \method consistently improves motion dynamics and effectively mitigates error accumulation during long-horizon generation. Fig.~\ref{fig:teaser} further shows that our method reduces long-rollout degradation while preserving the original few-step AR inference path.

 



%% file: section/related_work.tex
\section{Related Work}

\paragraph{Video diffusion foundation models.}
Recent progress in video generation has been largely built upon diffusion and flow-matching generative paradigms, which synthesize data through iterative denoising or continuous transport from noise to data~\citep{song2020denoising,flowmatching2023,ho2020denoising}. Combined with scalable diffusion transformer architectures~\cite{Peebles2022DiT}, these techniques have substantially improved text-to-video fidelity, motion quality, prompt following, and temporal coherence. Representative systems such as CogVideoX, HunyuanVideo, Wan, and LongCat-Video further scale video diffusion with stronger transformer backbones, latent video autoencoders, and large training corpora~\citep{cogvideox2024,hunyuanvideo2024,wan2025,longcatvideo2025,videoworldsimulators2024,opensora}. These models typically denoise a temporal window with bidirectional attention, allowing rich interactions among frames and leading to strong offline generation quality. However, their full-window denoising interface and multi-step sampling procedure are less suitable for real-time or interactive scenarios, where frames or chunks need to be emitted sequentially before the entire video is synthesized. This motivates causal autoregressive video generation, which preserves the generative strength of diffusion models while adapting them to streaming inference.

\paragraph{Autoregressive video generation and forcing-based training.}
Autoregressive (AR) video models generate videos sequentially by conditioning each new frame or chunk on previously generated content, often using transformer KV caches to efficiently reuse historical context. Early causal video diffusion methods commonly follow teacher-forcing or diffusion-forcing paradigms, where the model learns to denoise future frames conditioned on clean or partially noised context frames~\citep{jin2025pyramidal,diffusionforcing2024,chen2025skyreelsv2infinitelengthfilmgenerative}. To make AR diffusion practical for real-time generation, CausVid distills a bidirectional video DiT into a causal student with DMD-style few-step distillation~\citep{causvid2024,dmd2023}. Self-Forcing further reduces exposure bias by performing autoregressive self-rollout during training, allowing the model to condition on its own generated frames and rolling KV cache~\citep{selfforcing2025}. Following this direction, Self-Forcing++~\citep{selfforcingpp2025} extends self-rollout to minute-scale horizons; Causal Forcing improves causal AR distillation through a better initialization strategy~\citep{zhu2026causal}; Reward Forcing introduces reward feedback and EMA attention sinks to improve optimization and inference performance~\citep{rewardforcing2026}; and LongLive develops chunk-level AR generation with KV recache and streaming long tuning for interactive long videos~\citep{longlive2025}. Other works further improve cache management, long-context extrapolation, attention sinks, or inference-time stabilization for open-ended AR generation~\citep{rollingsink2026,rollingforcing2025,packforcing2026,deepforcing2025,yesiltepe2026infinity}. These methods improve the efficiency and stability of causal video generation, but their supervision is still primarily based on short-clip teachers, rollout-level objectives, reward signals, or cache heuristics. More recently, \citet{cai2026mmm} also explores real long-video data as supervision by introducing a two-head DiT design; however, it targets a non-causal long-video training paradigm rather than cache-aware post-training for causal few-step AR video generation. In contrast, \method focuses on post-training an already efficient few-step AR video model by providing dense denoising-level supervision on the student's own inference-time cache states.

\paragraph{On-policy self-distillation.}
On-policy self-distillation (OPSD) has recently emerged as a way to improve models on the trajectories they actually visit. In autoregressive large language models, the same model can act as both student and teacher under different contexts: the student samples from its current policy, while the teacher distribution is obtained by conditioning the model on richer in-context information~\citep{selfdistilledreasoner2026,sdft2026,sdpo2026,sdzero2026}. This idea has also been extended to step-distilled diffusion and flow models. D-OPSD applies OPSD to text-to-image generation by conditioning the teacher on both the text prompt and its paired real image, enabling supervised tuning along the student's own few-step rollouts without sacrificing few-step inference capability~\citep{dopsd2026}. Related OPD-style methods further study trajectory-level distillation for diffusion and flow models~\citep{flowopd2026,diffusionopd2026,anyflow2026}. \method follows this context-enhanced self-distillation principle, but extends it to the AR video setting where the on-policy state is not only the noisy latent and denoising timestep, but also the evolving KV cache written by previously generated chunks. By using real long videos as privileged temporal context, \method constructs a cleaner teacher distribution and provides cache-aware dense supervision for long-horizon few-step AR video generation.

%% file: section/preliminaries.tex
\section{Preliminaries}

\subsection{Few-Step Autoregressive Video Generation}

We consider a causal autoregressive (AR) video diffusion model that generates a video as a sequence of latent chunks. Let $x_{1:N}=\{x_1,\ldots,x_N\}$ denote a video divided into $N$ chunks, where each chunk may contain one or more latent frames, and let $c$ denote the text prompt or other conditioning signal. An AR video generator factorizes the video distribution as
\[
p_\theta(x_{1:N}\mid c)
=
\prod_{i=1}^{N} p_\theta(x_i \mid x_{<i}, c).
\]
In causal video diffusion transformers, the historical context $x_{<i}$ is implemented through transformer key-value (KV) caches, so previous chunks do not need to be recomputed from scratch at every generation step. We denote the cache state available before generating chunk $i$ as $h_i$. During inference, the model generates chunk $i$ conditioned on $h_i$, and then appends the generated chunk into the cache for future chunks.

For a few-step AR video generation model, each chunk is generated by a small number of denoising steps. Starting from Gaussian noise $z_{i,K}$, the model denoises it over a fixed timestep schedule $\{t_K,\ldots,t_1\}$. Let $f_\theta$ denote the causal video DiT velocity predictor. At denoising step $k$, the model predicts a velocity field
\[
\hat{v}_{i,k}
=
f_\theta(z_{i,k}, t_k, c, h_i),
\]
where $z_{i,k}$ is the current noisy latent of chunk $i$, $t_k$ is the denoising timestep, and $h_i$ is the historical KV-cache state. The predicted velocity $\hat{v}_{i,k}$ is not the next latent itself; instead, it specifies the denoising direction used by the numerical solver. Given this velocity, a solver transition $\Phi$ updates the latent state:
\[
z_{i,k-1}
=
\Phi(z_{i,k}, t_k, t_{k-1}, \hat{v}_{i,k}),
\qquad k=K,\ldots,1.
\]
After the final denoising step, the clean latent chunk is obtained as
\[
\hat{x}_i = z_{i,0}.
\]

The generated chunk is then written into the KV cache. We use $\operatorname{KV}_{\theta}(\hat{x}_i)$ to denote the KV entries obtained from the generated clean chunk, and use $\oplus$ to denote appending new KV entries to the existing cache. The cache update can be written as
\[
h_{i+1}
=
h_i \oplus \operatorname{KV}_{\theta}(\hat{x}_i).
\]
Thus, future chunks depend recursively on previously generated chunks through the evolving KV cache.

Few-step AR video generators are commonly obtained by combining causal video modeling with distribution matching distillation (DMD) or related few-step distillation objectives~\citep{dmd2023,decoupleddmd2025}. DMD compresses a strong but slow diffusion model into an efficient one-step or few-step generator by matching the student distribution to a teacher or data distribution. In AR video generation, methods such as CausVid use DMD-style distillation to adapt a bidirectional video diffusion teacher into a causal few-step student~\citep{causvid2024}, while Self-Forcing further performs autoregressive self-rollout during training so that the model conditions on its own generated frames and rolling KV cache~\citep{selfforcing2025}. These techniques provide the efficient few-step AR generators that serve as the starting point for our post-training framework.

\subsection{On-Policy Self-Distillation}
 
On-policy self-distillation (OPSD) has recently been studied in autoregressive language models as a way to improve a model on the trajectories it actually samples~\cite{selfdistilledreasoner2026,sdft2026}. Given an input query $q$, the model first acts as a student and samples an output trajectory from its current policy:
\begin{equation}
\hat{o}_{1:T} \sim \pi_\theta(\cdot \mid q),
\end{equation}
where $\hat{o}_{1:T}$ denotes the sampled sequence. The same model, or an exponential-moving-average copy of it, then acts as a teacher under a stronger context. Let $r$ denote additional in-context information, such as demonstrations, intermediate reasoning, or other privileged supervision. At each token position $m$, the student predicts the next-token distribution conditioned on the query and its own sampled prefix $\hat{o}_{<m}$, while the teacher predicts under the same sampled prefix but with the additional context $r$:
\begin{equation}
\pi_\theta(\cdot \mid q,\hat{o}_{<m}),
\qquad
\pi_{\bar{\theta}}(\cdot \mid q,r,\hat{o}_{<m}),
\end{equation}
where $\bar{\theta}$ denotes the teacher parameters. OPSD then optimizes the student by matching these two distributions along the student-sampled trajectory:
\begin{equation}
\mathcal{L}_{\mathrm{OPSD}}
=
\mathbb{E}_{\hat{o}_{1:T}\sim \pi_\theta(\cdot\mid q)}
\left[
\sum_{m=1}^{T}
D\left(
\pi_{\bar{\theta}}(\cdot \mid q,r,\hat{o}_{<m})
\;\|\;
\pi_{\theta}(\cdot \mid q,\hat{o}_{<m})
\right)
\right],
\end{equation}
where $D(\cdot\|\cdot)$ denotes a distillation divergence. The key idea is that the rollout prefix $\hat{o}_{<m}$ comes from the student itself, so the trajectory remains on-policy, while the teacher provides a stronger target by using the additional context $r$.

D-OPSD extends this idea to step-distilled text-to-image diffusion models~\citep{dopsd2026}. In D-OPSD, the student follows its own few-step denoising trajectory conditioned on the text prompt, while the teacher is conditioned on richer multimodal context constructed from the text prompt and its paired real image. The student is then supervised along its own rollout states, preserving the original few-step inference behavior. Our goal is to instantiate this principle in few-step AR video generation. Compared with image diffusion, the on-policy state in AR video is more complex: it includes not only the current noisy latent and denoising timestep, but also the temporal KV cache produced by previously generated chunks. In the following section, we introduce \method, which uses real long videos as privileged temporal context to construct a cleaner teacher distribution, while keeping the student rollout and cache dynamics aligned with inference.

%% file: section/method.tex
\section{Method}

\begin{figure}[t]
\centering
\includegraphics[width=\linewidth]{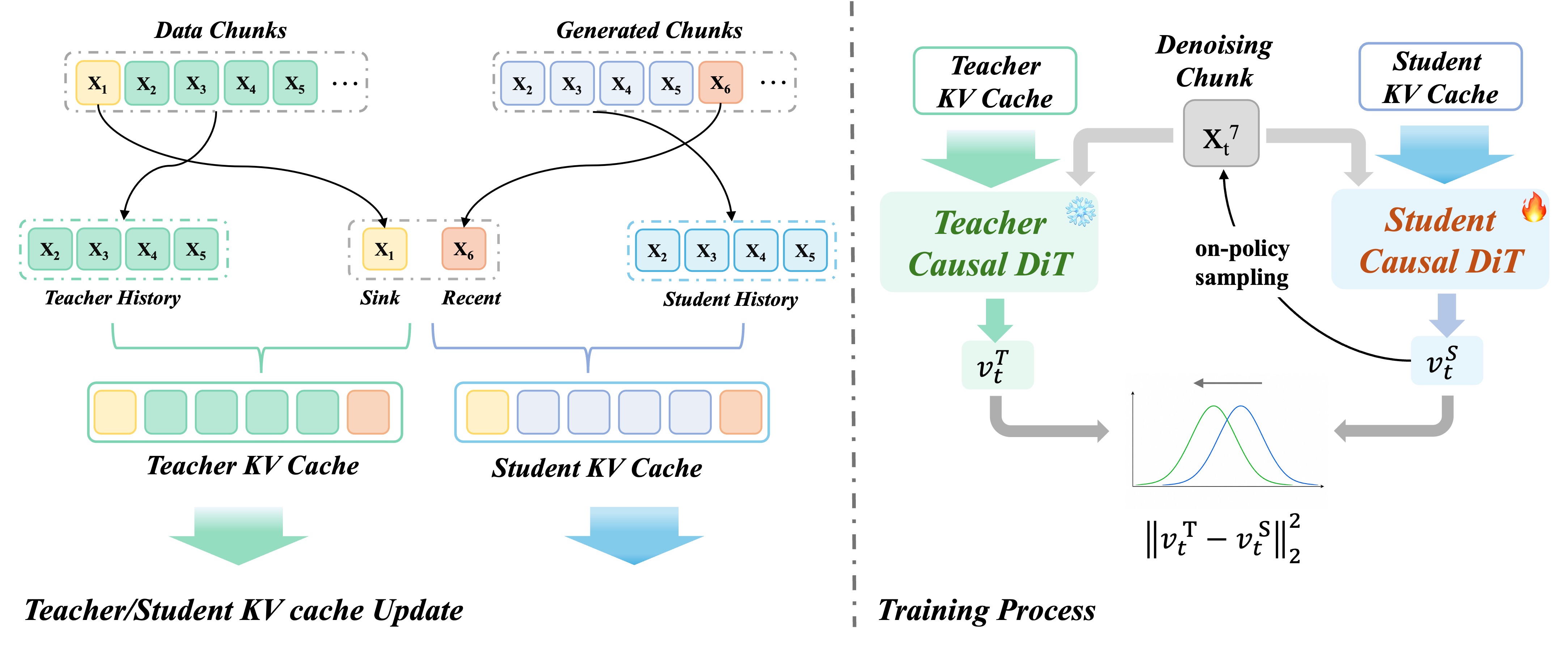}
\caption{
\method post-trains a few-step autoregressive video generator with cache-aware on-policy self-distillation. The student follows the exact inference-time rollout and writes its own generated chunks into the KV cache. The teacher is evaluated at the same student-visited denoising states, but uses an AR-consistent real-video cache, where older history is replaced by real-video chunks while the most recent chunk remains student-generated.
}
\label{fig:method_overview}
\end{figure}

\subsection{Overview}

We propose \method, a cache-aware on-policy self-distillation framework for post-training few-step causal AR video generators. Given a long training video divided into latent chunks $x^{\mathrm{data}}_{1:N}$, we use the first chunk $x^{\mathrm{data}}_1$ as a shared real-video prefix for both student and teacher. This prefix initializes the KV cache and anchors the scene, but does not participate in generation or loss computation. Starting from chunk $i=2$, the student follows the original inference-time rollout: it denoises each chunk with the fixed few-step sampler, writes its own generated chunk into the KV cache, and continues from this self-generated temporal state.

As shown in Fig.~\ref{fig:method_overview}, the key idea is to keep the student rollout fully on-policy while constructing a cleaner teacher distribution from real long-video context. For each generated chunk, the teacher is evaluated at the same student-visited noisy latents and denoising timesteps, but with a different temporal cache: older history is replaced by real-video chunks, while the most recent chunk remains student-generated to preserve autoregressive continuation. For example, at chunk $i=7$, the student conditions on ${x^{\mathrm{data}}_1,\hat{x}^s_2,\hat{x}^s_3,\hat{x}^s_4,\hat{x}^s_5,\hat{x}^s_6}$, whereas the teacher conditions on ${x^{\mathrm{data}}_1,x^{\mathrm{data}}_2,x^{\mathrm{data}}_3,x^{\mathrm{data}}_4,x^{\mathrm{data}}_5,\hat{x}^s_6}$. This design provides dense denoising-level corrective targets on the student's own rollout states without changing the original few-step sampling path. We next describe the student on-policy rollout, the construction of the real-video teacher cache, and the dense distillation objective.

 \subsection{Student On-Policy Rollout}

The student branch defines the on-policy rollout states used for training. We first initialize the student cache with the real first chunk $x^{\mathrm{data}}_1$. Specifically, $x^{\mathrm{data}}_1$ is passed through the causal DiT to obtain its KV entries:
\begin{equation}
    h_2^s = \operatorname{KV}_{\theta}(x^{\mathrm{data}}_1),
    \label{eq:student_init_cache}
\end{equation}
where $\operatorname{KV}_{\theta}(\cdot)$ denotes the operation that converts a clean chunk into the corresponding cached key-value entries. This first chunk serves only as a shared context and is not generated by the student.

Starting from chunk $i=2$, the student follows the exact inference-time procedure. Given the current student cache $h_i^s$, the student starts from Gaussian noise $z^s_{i,K}$ and denoises it with the fixed few-step sampler. At denoising step $k$, the student predicts a velocity field
\begin{equation}
    \hat{v}^s_{i,k}
    =
    f_{\theta}(z^s_{i,k}, t_k, c, h_i^s),
    \label{eq:student_velocity}
\end{equation}
where $z^s_{i,k}$ is the student noisy latent, $t_k$ is the denoising timestep, and $c$ is the text condition. The solver then updates the latent state by
\begin{equation}
    z^s_{i,k-1}
    =
    \operatorname{sg}\left(
    \Phi(z^s_{i,k}, t_k, t_{k-1}, \hat{v}^s_{i,k})
    \right),
    \qquad k=K,\ldots,1.
    \label{eq:student_solver}
\end{equation}
Here $\operatorname{sg}(\cdot)$ denotes stop-gradient through the rollout transition, which prevents losses on later chunks from backpropagating through the entire autoregressive history. During rollout, we record all student states $\{z^s_{i,k}, \hat{v}^s_{i,k}\}_{k=1}^{K}$ for teacher evaluation and distillation.

After the final denoising step, the generated chunk is
\begin{equation}
    \hat{x}^s_i = z^s_{i,0}.
\end{equation}
It is then written into the student cache using the same cache update rule as inference:
\begin{equation}
    h_{i+1}^s
    =
    h_i^s \oplus \operatorname{KV}_{\theta}(\hat{x}^s_i),
    \label{eq:student_cache_update}
\end{equation}
where $\oplus$ denotes appending the new KV entries to the existing cache.

Equivalently, before generating chunk $i$, the student cache contains the real first chunk followed by all previously generated student chunks:
\begin{equation}
    h_i^s
    \equiv
    \operatorname{KV}_{\theta}(x^{\mathrm{data}}_1)
    \oplus
    \operatorname{KV}_{\theta}(\hat{x}^s_2)
    \oplus
    \cdots
    \oplus
    \operatorname{KV}_{\theta}(\hat{x}^s_{i-1}).
    \label{eq:student_cache}
\end{equation}
Therefore, every supervised student prediction is conditioned on the same type of self-generated KV-cache state that the model will encounter during deployment.

\subsection{AR-Consistent Real-Video Teacher Cache}

The teacher branch provides cleaner corrective targets while preserving autoregressive continuation. Let $\bar{\theta}$ denote the teacher parameters. In our LoRA post-training setting, the base video generator is frozen, the student LoRA is trainable, and the teacher LoRA is maintained as an exponential-moving-average (EMA) copy of the student LoRA. The teacher is used only to produce stop-gradient targets.

For chunk $i$, the teacher is evaluated at the same student-visited noisy latent $z^s_{i,k}$ and timestep $t_k$, but with a different temporal cache $h_i^t$:
\begin{equation}
\hat{v}^t_{i,k}
=
f_{\bar{\theta}}(z^s_{i,k}, t_k, c, h_i^t).
\label{eq:teacher_velocity}
\end{equation}
Importantly, the teacher does not sample its own denoising trajectory. The denoising state remains on-policy because $z^s_{i,k}$ comes from the student rollout, while the teacher cache provides a cleaner temporal context for constructing the target velocity. In other words, the student trajectory determines where supervision is applied, and the teacher cache determines the corrective direction.

The teacher cache is constructed from real long-video chunks with an autoregressive consistency constraint. Before supervising chunk $i$, the student cache contains the real first chunk followed by previously generated student chunks:
\begin{equation}
h_i^s
\equiv
\operatorname{KV}_{\theta}(x^{\mathrm{data}}_1)
\oplus
\operatorname{KV}_{\theta}(\hat{x}^s_2)
\oplus
\cdots
\oplus
\operatorname{KV}_{\theta}(\hat{x}^s_{i-1}).
\label{eq:student_cache_conceptual}
\end{equation}
In contrast, the teacher cache replaces the older generated history with the corresponding real-video chunks, while keeping the most recent student-generated chunk:
\begin{equation}
h_i^t
\equiv
\operatorname{KV}_{\bar{\theta}}(x^{\mathrm{data}}_1)
\oplus
\operatorname{KV}_{\bar{\theta}}(x^{\mathrm{data}}_2)
\oplus
\cdots
\oplus
\operatorname{KV}_{\bar{\theta}}(x^{\mathrm{data}}_{i-2})
\oplus
\operatorname{KV}_{\bar{\theta}}(\hat{x}^s_{i-1}).
\label{eq:teacher_cache}
\end{equation}
For example, at chunk $i=7$, the two caches are
\begin{equation}
\begin{aligned}
h_7^s &: [x^{\mathrm{data}}_1,\hat{x}^s_2,\hat{x}^s_3,\hat{x}^s_4,\hat{x}^s_5,\hat{x}^s_6],\\
h_7^t &: [x^{\mathrm{data}}_1,x^{\mathrm{data}}_2,x^{\mathrm{data}}_3,x^{\mathrm{data}}_4,x^{\mathrm{data}}_5,\hat{x}^s_6].
\end{aligned}
\end{equation}

This cache policy balances two goals. Replacing older history with real-video chunks reduces accumulated error contamination in the teacher context and provides a cleaner long-range temporal state. Keeping the most recent student-generated chunk prevents the teacher from becoming a fully teacher-forced oracle: the teacher still predicts how to continue from the student's latest generated state, which better matches autoregressive deployment. Both student and teacher use the same attention-sink cache mechanism with rolling relative RoPE. The sink positions are updated consistently as the rollout grows, so the difference between student and teacher predictions mainly comes from cache content rather than inconsistent positional indexing.

\subsection{Dense Denoising-Level Objective}

\method provides supervision at the fixed denoising steps used by the few-step sampler. For each supervised chunk-step pair $(i,k)$, the student prediction $\hat{v}^s_{i,k}$ and the teacher target $\hat{v}^t_{i,k}$ are evaluated at the same student-visited noisy latent $z^s_{i,k}$ and timestep $t_k$, but under different temporal caches. We use a pure velocity matching objective:
\begin{equation}
\mathcal{L}_{\mathrm{OPSD\text{-}V}}
=
\frac{1}{|\mathcal{S}|}
\sum_{(i,k)\in\mathcal{S}}
\left\|
\hat{v}^s_{i,k}
-
\operatorname{sg}(\hat{v}^t_{i,k})
\right\|_2^2 .
\label{eq:opsdv_loss}
\end{equation}
where $\operatorname{sg}(\cdot)$ stops gradients through the teacher prediction, and $\mathcal{S}$ denotes the set of supervised chunk-step pairs.

In our implementation, the underlying AR generator uses a four-step sampler, and we supervise all four denoising steps. We apply a rollout warm-up by excluding the first $M=7$ chunks from the loss:
\begin{equation}
    \mathcal{S}
    =
    \left\{
    (i,k)
    \,\middle|\,
    i > M,\;
    k=1,\ldots,K
    \right\},
    \qquad M=7,\; K=4.
    \label{eq:supervised_set}
\end{equation}
The warm-up chunks are still generated autoregressively and written into the student cache, but they do not contribute to the distillation loss. We set $M=7$ because the Wan-based AR models used in our experiments are originally trained within a local window of seven chunks, corresponding to $3\times7=21$ latent frames. Starting the loss after this local horizon encourages supervision to focus on long-rollout states where accumulated cache degradation begins to appear. The loss is then applied to all subsequent chunks in the same contiguous long-video rollout.


\subsection{Training Procedure}
\label{sec:training_procedure}

We summarize the training procedure of \method in Algorithm~\ref{alg:opsdv}. For each long training video, both branches start from the same real first chunk. The student then performs autoregressive self-rollout to define the on-policy denoising states. After the rollout warm-up, the teacher cache is constructed for each supervised chunk with real-video older history and the most recent student-generated chunk, as defined in Eq.~\eqref{eq:teacher_cache}. The EMA teacher is evaluated at the same student-visited noisy latents, and the student is optimized with dense velocity matching on these supervised chunks.

\paragraph{Memory-efficient truncated backpropagation.}
A naive implementation would retain the computation graphs of all supervised chunk-step pairs until the complete long-video rollout is finished, causing activation memory to grow with both rollout length and the number of denoising steps. This is unnecessary in our formulation because the solver transition in Eq.~\eqref{eq:student_solver} is stop-gradient and all KV-cache writes are performed without gradients. Consequently, the gradient of Eq.~\eqref{eq:opsdv_loss} decomposes over supervised pairs:
\begin{equation}
\nabla_{\theta}\mathcal{L}_{\mathrm{OPSD\text{-}V}}
=
\sum_{(i,k)\in\mathcal{S}}
\nabla_{\theta}
\frac{\ell_{i,k}}{|\mathcal{S}|},
\qquad
\ell_{i,k}
=
\left\|
\hat{v}^s_{i,k}-\operatorname{sg}(\hat{v}^t_{i,k})
\right\|_2^2.
\label{eq:online_backward}
\end{equation}
We therefore backpropagate each normalized term $\ell_{i,k}/|\mathcal{S}|$ immediately after its student forward pass and accumulate parameter gradients without updating the model. The current activation graph is then released before proceeding to the next denoising step. Teacher predictions for supervised chunks, warm-up student predictions, denoising transitions, and cache updates are all evaluated without gradients. The optimizer is stepped only once after the complete rollout, followed by the EMA teacher update. This online accumulation is mathematically equivalent to backpropagating the summed objective, while retaining at most one gradient-enabled student forward graph at a time. Thus, activation memory does not grow with the number of supervised chunk-step pairs; only the detached KV caches and accumulated parameter gradients persist across the rollout.

\begin{algorithm}[t]
\caption{Memory-Efficient Training Procedure of \method}
\label{alg:opsdv}
\begin{algorithmic}[1]
\REQUIRE Long-video dataset $\mathcal{D}$, student model $f_{\theta}$, EMA teacher $f_{\bar{\theta}}$, denoising steps $K$, warm-up chunks $M$
\WHILE{not converged}
    \STATE Sample a contiguous long video and condition $(x^{\mathrm{data}}_{1:N}, c) \sim \mathcal{D}$.
    \STATE Initialize the detached student cache from the shared prefix $x^{\mathrm{data}}_1$.
    \STATE Initialize the accumulated parameter gradient: $g \leftarrow 0$.
    \FOR{$i=2$ to $N$}
        \STATE Sample initial noise $z^s_{i,K} \sim \mathcal{N}(0,I)$.
        \IF{$i > M$}
            \STATE Construct $h_i^t$ with real older history and the latest student-generated chunk, as in Eq.~(\ref{eq:teacher_cache}).
        \ENDIF
        \FOR{$k=K$ to $1$}
            \IF{$i > M$}
                \STATE Predict teacher target without gradients: $\hat{v}^t_{i,k} \leftarrow f_{\bar{\theta}}(z^s_{i,k}, t_k, c, h^t_i)$.
                \STATE Predict student velocity with gradients: $\hat{v}^s_{i,k} \leftarrow f_{\theta}(z^s_{i,k}, t_k, c, h^s_i)$.
                \STATE Compute $\ell_{i,k} \leftarrow \left\|\hat{v}^s_{i,k}-\operatorname{sg}(\hat{v}^t_{i,k})\right\|_2^2$.
                \STATE Backpropagate immediately: $g \leftarrow g + \nabla_{\theta}\!\left(\ell_{i,k}/|\mathcal{S}|\right)$.
                \STATE Release the current activation graph and detach all cache tensors.
            \ELSE
                \STATE Predict student velocity without gradients: $\hat{v}^s_{i,k} \leftarrow f_{\theta}(z^s_{i,k}, t_k, c, h^s_i)$.
            \ENDIF
            \STATE Update student latent without gradients: $z^s_{i,k-1} \leftarrow \operatorname{sg}\!\left(\Phi(z^s_{i,k}, t_k, t_{k-1}, \hat{v}^s_{i,k})\right)$.
        \ENDFOR
        \STATE Obtain generated chunk: $\hat{x}^s_i \leftarrow z^s_{i,0}$.
        \STATE Append $\hat{x}^s_i$ to the detached student cache.
    \ENDFOR
    \STATE Clip $g$ and update the student LoRA parameters $\theta$ once.
    \STATE Update the teacher LoRA parameters $\bar{\theta}$ with EMA.
\ENDWHILE
\end{algorithmic}
\end{algorithm}

In our experiments, $K=4$ and all denoising steps are supervised. We set $M=7$, so the first seven chunks are still generated and written into the student cache, but they do not contribute to the loss. This warm-up lets the student enter long-rollout cache states before dense distillation begins. Although all four denoising steps receive supervision, their activation graphs are processed and released one at a time according to Eq.~\eqref{eq:online_backward}. At inference time, the teacher branch and real-video cache are removed; the model uses the original few-step AR sampling path.

%% file: section/experiments.tex
\section{Experiments}
\begin{figure}[!t]
\centering
\includegraphics[width=0.96\linewidth,height=0.74\textheight,keepaspectratio]{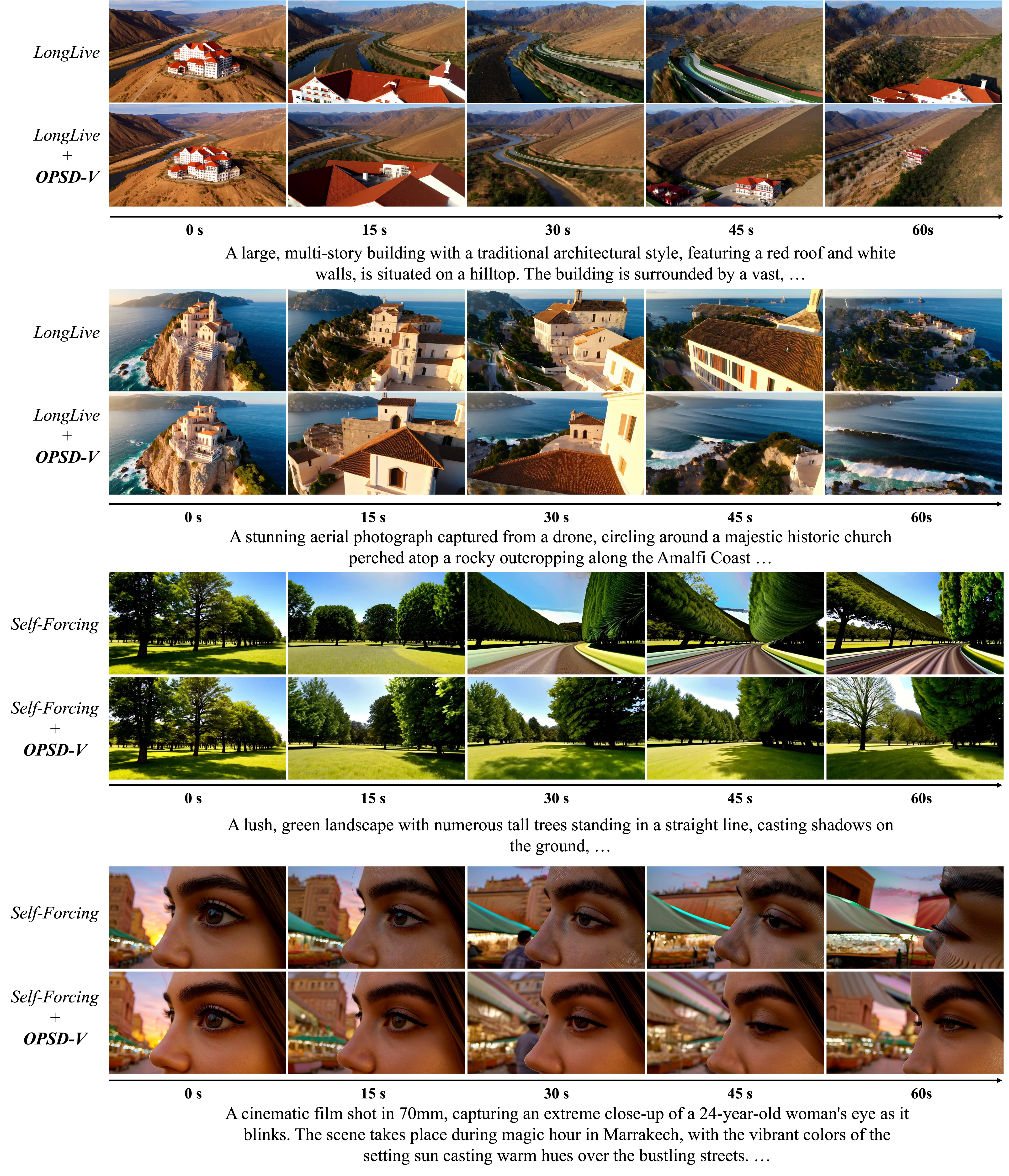}
\caption{
Qualitative comparison on long autoregressive video generation. The first two examples are based on LongLive~\cite{longlive2025}, and the last two examples are based on Self-Forcing~\cite{selfforcing2025}. For each example, the first row shows the result generated by the original base model, and the second row shows the result after applying \method post-training. Each pair uses the same prompt, the same random seed, the same 4-step sampler, and the same attention-sink inference setting. Compared with the base models, \method better preserves motion dynamics and visual quality during long rollout, reducing common AR degradation artifacts such as weakened motion, blur, temporal flickering, and semantic drift in later chunks.
}
\label{fig:results_all}
\end{figure}
\subsection{Experimental Setup}

\paragraph{Training data.}
We construct a small customized long-video dataset for \method post-training. The dataset contains 3,800 videos, each approximately one minute in length and processed at 480p resolution. The videos cover diverse content, including natural landscapes, large camera motions, and human-centered scenes. We apply a lightweight filtering process based on optical flow and other basic quality cues to remove low-motion, unstable, or low-quality samples. All videos are encoded into latent chunks using the Wan2.1 VAE~\cite{wan2025}. Compared with short-clip training data, these long videos provide richer temporal context and are therefore suitable for constructing the real-video teacher cache in our framework. During training, we use the first chunk as the shared real-video prefix for both student and teacher, and perform autoregressive rollout on subsequent chunks.

\paragraph{Base models.}
We evaluate \method on two representative few-step AR video generators: Self-Forcing~\citep{selfforcing2025} and LongLive~\citep{longlive2025}. Both models are built upon the Wan2.1-T2V-1.3B backbone~\cite{wan2025}. For Self-Forcing, we start from its official base checkpoint and train a LoRA adapter for our post-training. For LongLive, we use its official base model and released LoRA checkpoint, and continue post-training the LoRA adapter with \method. Evaluating on both models allows us to test whether \method can serve as a general post-training refinement method for different few-step AR video generators.

\paragraph{Post-training setting.}
The student branch uses the trainable LoRA adapter, while the teacher branch uses an exponential-moving-average (EMA) copy of the student LoRA with a decay rate of 0.9999. The teacher is only used during training and is removed at inference time. We use the original four-step sampler of the base AR video models and supervise all four denoising steps with the velocity matching loss in Eq.~\eqref{eq:opsdv_loss}. No additional DMD, reward, or adversarial loss is used. We train each model for 200 iterations on 24 H800 GPUs with a per-GPU batch size of 1. Training uses BF16 mixed precision, activation checkpointing, and fully sharded data parallelism. Together with the immediate per-step backward strategy described in Sec.~\ref{sec:training_procedure}, these choices enable long-rollout post-training without retaining the full rollout computation graph.

\paragraph{Rollout and cache construction.}
Each training sample is a contiguous long-video segment. For each rollout, we use 180 latent frames in total, divided into 60 chunks, with each chunk containing 3 latent frames. Both student and teacher are initialized from the same real first chunk. The student then follows the exact inference-time rollout, writing each generated chunk into its KV cache. The teacher cache is constructed with real-video chunks for older history and the most recent student-generated chunk for autoregressive continuation. We exclude the first $M=7$ chunks from the distillation loss, corresponding to the original local training horizon of the Wan-based AR models. These warm-up chunks are still generated and written into the student cache, but only subsequent chunks contribute to the loss.

\paragraph{Evaluation protocol.}
We evaluate \method both qualitatively and quantitatively on one-minute video generation at 16 FPS. For qualitative comparison, we use the same prompt, random seed, sampler, and attention-sink inference setting for each pair of baseline and \method post-trained models. This allows us to directly compare the effect of our post-training objective under identical inference conditions. For quantitative evaluation, we use 240 prompts in total: the first 120 prompts from MovieGenBench~\citep{polyak2024movie} and 120 internal prompts, which we refer to as MeiBench. Each method generates one video per prompt, and the resulting 240 videos are evaluated with VBenchLong~\citep{huang2025vbench++}. We report averaged results over MovieGenBench and MeiBench. All compared models use the same 4-step AR inference path and the same inference-time KV-cache mechanism as their corresponding base models. Therefore, any improvement comes from post-training rather than additional inference computation.

\subsection{Qualitative Comparison}

Fig.~\ref{fig:results_all} compares the original few-step AR generators with their \method post-trained counterparts under identical inference settings. Across both LongLive and Self-Forcing examples, the base models can produce plausible early chunks, but visual quality tends to degrade as autoregressive rollout continues. Typical failure modes include weakened motion, accumulated blur, local artifacts, and semantic or structural drift in later frames. After applying \method, the generated videos better preserve scene structure and motion dynamics over the full one-minute horizon. For LongLive, \method maintains sharper background details and more stable camera evolution in long landscape sequences. For Self-Forcing, \method produces more coherent object motion and reduces late-stage artifacts around the subject and background. These improvements are obtained without changing the sampler, number of denoising steps, or inference-time KV-cache mechanism, suggesting that cache-aware on-policy supervision improves the model's behavior on the states it naturally visits during deployment.

\begin{table}[!t]
\centering
\caption{
Quantitative comparison on one-minute video generation. We generate 240 videos in total, including 120 prompts from MovieGenBench and 120 internal prompts from MeiBench, and evaluate all results with VBenchLong~\citep{huang2025vbench++}. All methods are based on the same Wan2.1-T2V-1.3B backbone, use the same 4-step autoregressive inference path, and adopt the same attention-sink cache mechanism. Higher scores are better. \method improves Quality Score and Dynamic Degree on both LongLive and Self-Forcing without increasing inference cost.
}
\label{tab:main_quantitative}
\renewcommand{\arraystretch}{1.12}
\setlength{\tabcolsep}{5pt}
\resizebox{\linewidth}{!}{
\begin{tabular}{lccccc}
\toprule
\rowcolor{headerblue}
Method 
& Params 
& NFE 
& Quality Score $\uparrow$
& Dynamic Degree $\uparrow$
& Semantic Score $\uparrow$ \\
\midrule
LongLive
& 1.3B 
& 4 
& 0.8138 
& 0.5012 
& \best{0.4911} \\

\rowcolor{oursblue}
LongLive + \method
& 1.3B 
& 4 
& \best{0.8242} 
& \best{0.5890} 
& 0.4904 \\
\midrule

Self-Forcing
& 1.3B 
& 4 
& 0.8259 
& 0.6218 
& \best{0.4897} \\

\rowcolor{oursblue}
Self-Forcing + \method
& 1.3B 
& 4 
& \best{0.8389} 
& \best{0.6570} 
& 0.4864 \\
\bottomrule
\end{tabular}
}
\end{table}

\subsection{Quantitative Comparison}
Tab.~\ref{tab:main_quantitative} reports the quantitative results averaged over MovieGenBench and MeiBench. Across both LongLive and Self-Forcing backbones, \method consistently improves Quality Score and Dynamic Degree while using the same model scale, 4-step inference path, and attention-sink cache mechanism. These results indicate that our cache-aware on-policy distillation improves long-video generation quality mainly by enhancing motion dynamics, rather than relying on extra inference computation. The Semantic Score remains comparable to the corresponding base models, with a slight decrease after post-training, suggesting a mild trade-off between more active motion generation and conservative semantic matching metrics.

\begin{figure}[!t]
\centering
\newcommand{\userstudybar}[5]{%
  \node[anchor=east,font=\sffamily\bfseries\scriptsize,inner sep=0pt] at (-0.25,#1) {\shortstack[r]{#2}};
  \fill[opsdgreen] (0,#1-0.20) rectangle ({#3/10},#1+0.20);
  \fill[studysame] ({#3/10},#1-0.20) rectangle ({(#3+#4)/10},#1+0.20);
  \fill[studybase] ({(#3+#4)/10},#1-0.20) rectangle (10,#1+0.20);
  \node[white,font=\sffamily\bfseries\tiny] at ({#3/20},#1) {#3\%};
  \node[white,font=\sffamily\bfseries\tiny] at ({(#3+#3+#4)/20},#1) {#4\%};
  \node[white,font=\sffamily\bfseries\tiny] at ({(#3+#4+100)/20},#1) {#5\%};
}
\begin{tikzpicture}[x=0.46cm,y=0.76cm]
\node[font=\sffamily\bfseries\small] at (5,3.25) {LongLive};
\userstudybar{2.55}{Motion\\Quality}{49}{36}{15}
\userstudybar{1.55}{Visual\\Quality}{42}{33}{25}
\userstudybar{0.55}{Overall}{54}{28}{18}
\begin{scope}[xshift=6.8cm]
\node[font=\sffamily\bfseries\small] at (5,3.25) {Self-Forcing};
\userstudybar{2.55}{Motion\\Quality}{66}{25}{9}
\userstudybar{1.55}{Visual\\Quality}{79}{12}{9}
\userstudybar{0.55}{Overall}{78}{12}{10}
\end{scope}
\node[font=\sffamily\footnotesize] at (12.4,-0.25) {
\textcolor{opsdgreen}{\rule{0.8em}{0.8em}}~\method\quad
\textcolor{studysame}{\rule{0.8em}{0.8em}}~Same\quad
\textcolor{studybase}{\rule{0.8em}{0.8em}}~Base
};
\end{tikzpicture}
\caption{
\textbf{User study preference results.}
Ten participants compare 20 paired videos each, consisting of 10 LongLive pairs and 10 Self-Forcing pairs. Each pair contains a base-model video and its \method post-trained counterpart under the same prompt. For each pair, participants choose Model A, Model B, or Same for overall preference, motion quality, and visual quality. Each panel shows percentages over 100 judgments per criterion.
}
\label{fig:user_study}
\end{figure}
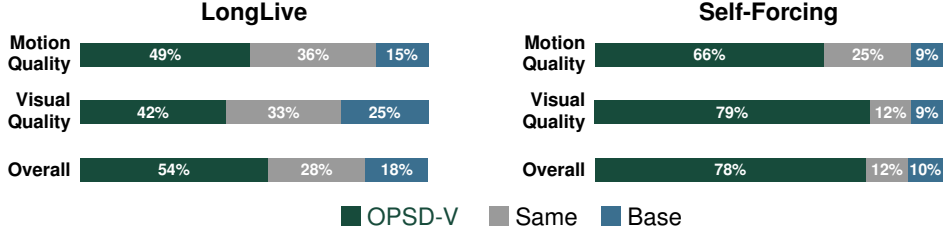

We further conduct a user study with 10 participants. Each participant compares 20 paired videos, including 10 LongLive pairs and 10 Self-Forcing pairs. Each pair contains one video generated by the base model and one generated after applying \method, shown together with the corresponding prompt. For each pair, participants answer three questions: \emph{Overall} preference considering all factors, \emph{Motion Quality} focusing on smoothness, naturalness, and absence of jitter or discontinuity, and \emph{Visual Quality} focusing on clarity, level of detail, and overall aesthetic quality. For each question, participants choose Model A, Model B, or Same when there is no perceptible difference. Fig.~\ref{fig:user_study} reports the results separately for the two backbones. On LongLive, \method is favored over the base model in overall and motion-quality judgments, while visual quality shows a larger fraction of ties and base-model preferences. On Self-Forcing, \method receives strong preference across all three criteria. Aggregated over both backbones, \method is preferred in 66.0\% of overall-quality comparisons, or 82.5\% after excluding ties. Users also favor \method for motion quality (57.5\%, 82.7\% excluding ties) and visual quality (60.5\%, 78.1\% excluding ties). Overall, the quantitative results and human preferences support \method as an effective post-training refinement strategy for few-step AR video generation.

\FloatBarrier

\subsection{Ablation Study}

We study two design choices that directly affect the stability of long-horizon post-training: the prediction space used for distillation and the trajectory on which the distillation targets are evaluated. Unless otherwise specified, all variants use the same training videos, four-step sampler, rollout length, cache construction, and LoRA post-training setup as our main method.

\begin{figure}[H]
\centering
\includegraphics[width=\linewidth]{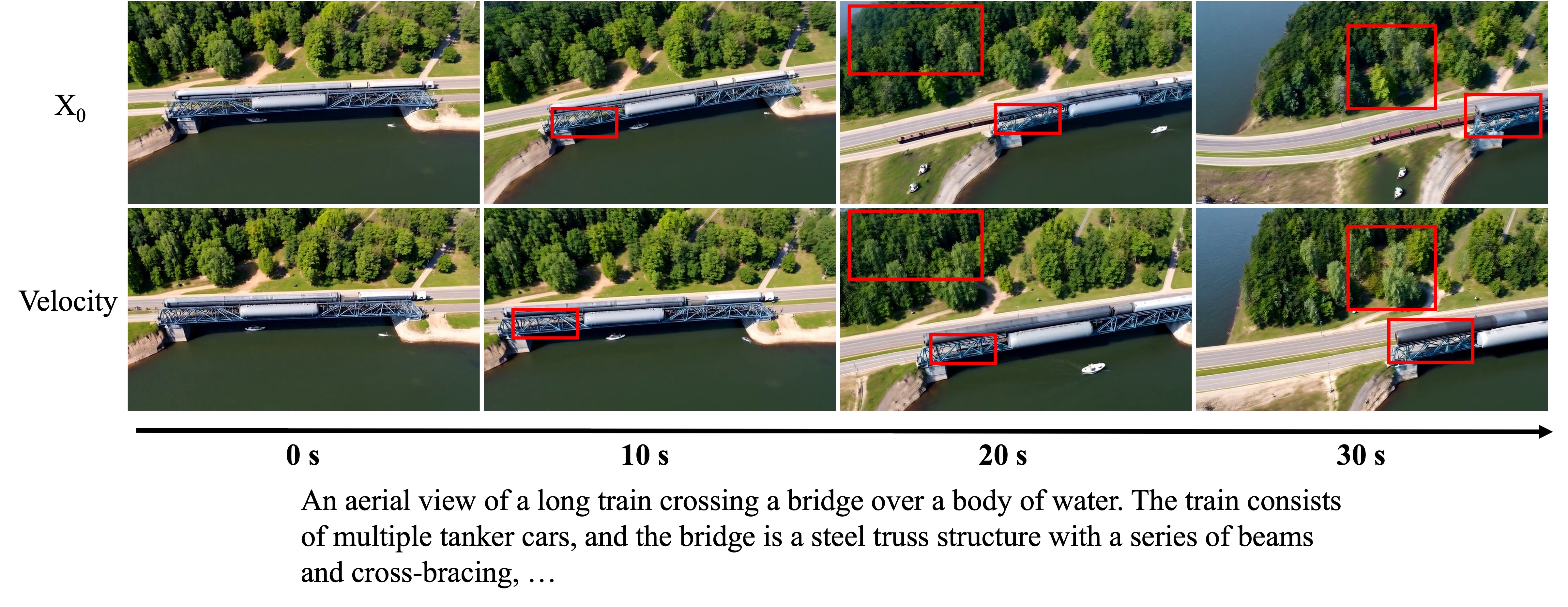}
\caption{
\textbf{Velocity versus clean-latent matching.}
We compare distillation in the predicted clean-latent ($x_0$) space and in the native velocity space over a one-minute rollout. Both variants remain plausible at the beginning, but $x_0$ matching progressively smooths fine structures and introduces geometric degradation in the bridge truss and foliage. Velocity matching better preserves these details at later timestamps. Red boxes highlight representative regions.
}
\label{fig:ablation_vx0}
\end{figure}

\paragraph{Velocity versus clean-latent matching.}
Our default objective directly matches the student and teacher velocity predictions, as defined in Eq.~\eqref{eq:opsdv_loss}. As an alternative, we convert both predictions into clean-latent estimates at every denoising step and optimize
\begin{equation}
\mathcal{L}_{x_0}
=
\frac{1}{|\mathcal{S}|}
\sum_{(i,k)\in\mathcal{S}}
\left\|
\hat{x}^{s}_{0,i,k}
-
\operatorname{sg}(\hat{x}^{t}_{0,i,k})
\right\|_2^2.
\end{equation}
Under the flow parameterization, converting a velocity prediction to $x_0$ introduces a timestep-dependent scale. Consequently, an $x_0$-space MSE reweights prediction errors across the four fixed denoising steps and places relatively greater emphasis on high-noise states. As shown in Fig.~\ref{fig:ablation_vx0}, this variant produces increasingly smooth details during long rollout, particularly in the bridge structure and tree boundaries. Direct velocity matching instead supervises the model in its native prediction space and better preserves high-frequency structure. We therefore use velocity matching for all main experiments.

\begin{figure}[H]
\centering
\includegraphics[width=\linewidth]{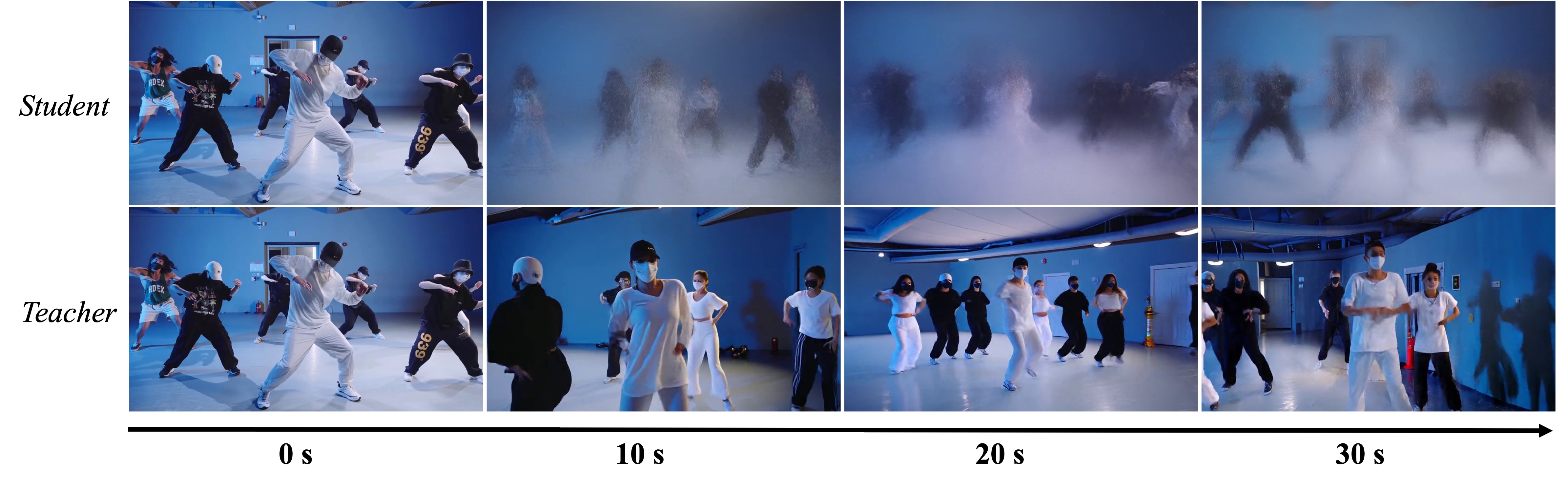}
\caption{
\textbf{Failure of teacher-trajectory supervision.}
When distillation states are sampled from the teacher's own denoising trajectory, the teacher rollout remains sharp and coherent (bottom), but the student becomes severely blurred under its own autoregressive rollout (top). This discrepancy reveals the off-policy state mismatch introduced by teacher-trajectory supervision.
}
\label{fig:ablation_student_teacher_trajectory}
\end{figure}

\paragraph{Student versus teacher trajectory.}
A central design choice in \method is that the teacher is evaluated at the noisy latents visited by the student, $z^s_{i,k}$, rather than sampling an independent denoising trajectory. We ablate this choice by constructing the distillation pairs from the teacher's own trajectory $z^t_{i,k}$. Although these teacher states yield visually cleaner targets, they are not the states that the student encounters when generating autoregressively at inference time. As a result, the student receives little supervision on how to recover from errors in its own denoising states, and the mismatch is repeatedly amplified after each generated chunk is written into the KV cache. Fig.~\ref{fig:ablation_student_teacher_trajectory} shows the resulting failure mode: the teacher trajectory remains clear, whereas the student rollout rapidly loses spatial detail and becomes dominated by blur. This result confirms that the teacher should provide the corrective prediction, but the student must determine the states on which that prediction is evaluated. We therefore keep the denoising trajectory fully on-policy in our final design.

%% file: section/analysis.tex
\section{Analysis and Future Work}

Our work starts from a simple question: what should be used as the target distribution when post-training few-step autoregressive video generators? Existing DMD-style approaches usually rely on a short-clip diffusion model as the teacher distribution. While effective for accelerating generation, such a target is inherently limited by the temporal range and quality of the teacher model, making it difficult to provide reliable supervision for truly long autoregressive rollouts. A natural alternative is to directly exploit real long videos, which contain richer temporal structure and cleaner long-range dynamics.

However, incorporating real videos into few-step AR post-training is not straightforward. Direct teacher forcing would break the inference-time rollout distribution, while using real videos only as reconstruction targets would not correct the model on its own generated cache states. Our key observation is that on-policy self-distillation provides a suitable interface for this problem: real videos can be treated as privileged temporal context rather than direct output targets. In this way, the student still follows its own inference-time trajectory, while the teacher uses cleaner real-video cache states to provide dense denoising-level corrective supervision.

This perspective suggests a promising direction for future work. Since \method explicitly leverages real long-video data, its performance may further benefit from scaling the amount, diversity, and quality of training videos, as well as increasing training compute. Moreover, our current cache construction and loss design are only one possible instantiation of this idea. More effective teacher-cache policies, adaptive supervision schedules, or stronger context-enhanced teachers may further improve long-horizon generation. We hope that \method can serve as a useful baseline and provide insight for future research on data-driven post-training of causal autoregressive video models.




%% file: section/conclusion.tex
\section{Conclusion}

We presented \method, a cache-aware on-policy self-distillation framework for post-training few-step autoregressive video generators. Instead of relying on short-clip teacher targets or fully teacher-forced training, \method supervises the student on its own inference-time rollout states while using real long-video context to construct cleaner teacher caches. This provides dense denoising-level correction under self-generated KV-cache states, improving long-horizon generation behavior without changing the original few-step inference path. Experiments on Self-Forcing and LongLive show that \method improves video quality and motion dynamics while preserving the same model scale and inference cost. We hope this work provides a useful step toward data-driven post-training for causal long-video generation.
